\pdfoutput=1

\documentclass[letterpaper,journal]{IEEEtran}

\usepackage{amsmath,amsfonts}
\usepackage{algorithm} 
\usepackage{array}
\usepackage[caption=false,font=footnotesize]{subfig}
\usepackage{textcomp}
\usepackage{stfloats}
\usepackage{url}
\usepackage{verbatim}
\usepackage{graphicx}
\usepackage{cite}

\usepackage{booktabs}   
\usepackage{multirow}   
\usepackage{algpseudocode} 
\usepackage{pgfplots}   
\pgfplotsset{compat=1.18}
\usepackage{enumitem}   
\usepackage{amssymb}    
\usepackage{mathtools}  
\usepackage{microtype}  
\usepackage{amsthm}     

\usepackage[colorlinks=true, allcolors=blue]{hyperref}

\newcommand{\method}{Look-ahead Sync}

\theoremstyle{remark}

\graphicspath{{figures/}} 

\begin{document}

\title{A High-Capacity and Secure Disambiguation Algorithm for Neural Linguistic Steganography}

\author{%
    Yapei Feng\textsuperscript{1},%
    Feng Jiang\textsuperscript{1},%
    Shanhao Wu\textsuperscript{2},%
    and Hua Zhong\textsuperscript{1}%
    \thanks{\textsuperscript{1}The authors are with the School of Cyberspace, Hangzhou Dianzi University, Hangzhou 310018, P. R. China (email: fengyapei@hdu.edu.cn).}%
    \thanks{\textsuperscript{2}Shanhao Wu is with the Bridge and Wind Engineering Laboratory, Department of Civil Engineering, School of Engineering, University of Tokyo, Tokyo 113-8656, Japan.}%
}


\maketitle

\begin{abstract}
Neural linguistic steganography aims to embed information into natural text while preserving statistical undetectability. A fundamental challenge in this field stems from tokenization ambiguity in modern tokenizers, which can lead to catastrophic decoding failures. The recent method, SyncPool, addresses this ambiguity by employing a coarse-grained synchronization mechanism over groups of ambiguous candidates. However, SyncPool sacrifices embedding capacity, as it utilizes the entire Shannon entropy of an ambiguous group solely for synchronization rather than for payload embedding.
We propose a method named look-ahead Sync, which overcomes the capacity limitation of SyncPool while retaining its provable security guarantees. Our approach performs minimal synchronized sampling only on truly indistinguishable token sequences, while strategically preserving all other discernible paths to maximize embedding capacity. We provide theoretical proofs for the security of our method and analyze the gap between its achievable embedding capacity and the theoretical upper bound.
Experiments on English (using Llama 3) and Chinese (using Qwen 2.5) benchmarks show that our method consistently approaches the theoretical capacity upper bound and significantly outperforms SyncPool. The improvement in embedding rate exceeds 160\% in English and 25\% in Chinese, particularly in settings with larger candidate pools. This work represents a significant step toward practical high-capacity provably secure linguistic steganography.
\end{abstract}


\section{Introduction}
\label{sec:introduction}

\IEEEPARstart{L}{inguistic} steganography hides data in ordinary text to enable covert communication while concealing the existence of the message. The effectiveness of such systems is assessed along two often competing objectives, embedding capacity and statistical security\cite{zhang2021provably,ding2023discop,de2022perfectly}. Capacity, measured in bits per token (BPT), quantifies how much secret information a covertext can carry and is critical for practical utility\cite{ziegler2019neural,yang2018rnn}. Security measures resistance to detection.\cite{yang2020linguistic,yang2020vae,wang2023linguistic,yang2022linguistic} The strongest standard is zero Kullback--Leibler (KL) divergence, which requires the distribution of stegotext to equal that of covertext\cite{cachin1998information}. Under this standard, no statistical test of any power can distinguish the two\cite{cachin1998information}. The central challenge is therefore to maximize capacity without relaxing this criterion\cite{hopper2002provably}.

\begin{figure}[t]
    \centering
    \includegraphics[width=\columnwidth]{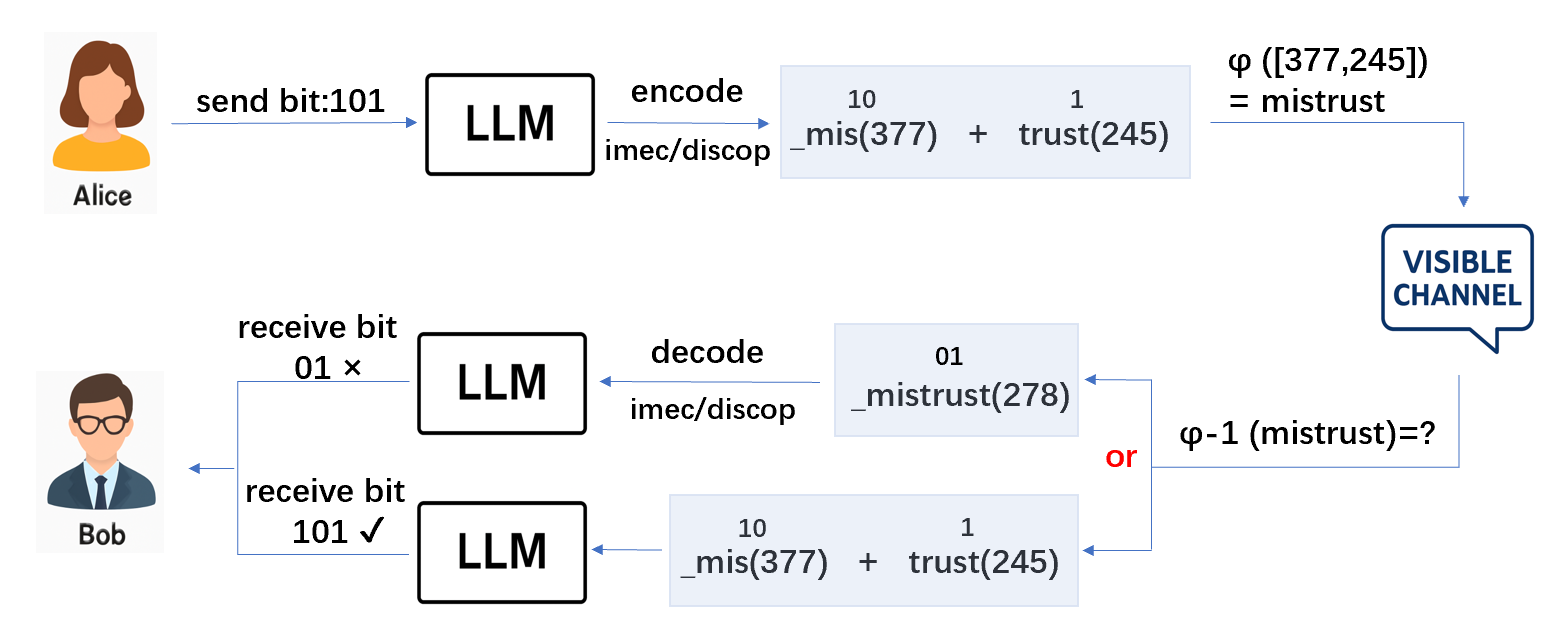}
    \caption{A toy example of tokenization ambiguity. The detokenizer $\varphi$ is not injective, so identical text (mistrust) can correspond to [278] or [377, 245]. If bits are embedded at the token level without resolving this ambiguity, the receiver may decode the wrong bits from the same visible string.}
    \label{fig:ambiguity}
\end{figure}

The advent of large language models (LLMs) provides fluent covertext\cite{vaswani2017attention,brown2020language,touvron2023llama}, yet their subword tokenization schemes, such as Byte Pair Encoding (BPE)~\cite{sennrich2015neural,kudo2018sentencepiece}, introduce the critical challenge of tokenization ambiguity, where a single visible string can correspond to multiple token sequences\cite{nozaki2022addressing}. For example, the string mistrust may be tokenized as [\_mistrust] or as [\_mis, \_trust] (see Fig.~\ref{fig:ambiguity}). In autoregressive generation, if the receiver reconstructs a different token path than the sender, the conditional distributions for subsequent steps become desynchronized, making the remaining payload unrecoverable and causing catastrophic decoding failure~\cite{qi2024provably}.

Addressing this ambiguity requires a dedicated disambiguation module. Early generative approaches by Nozaki and Murawaki~\cite{nozaki2022addressing} enforced a unique decoding path by pruning any token that is a prefix of another candidate. Although this guarantees decodability, pruning irreversibly alters the model's distribution. Variants based on a maximum-weight independent set (MWIS)~\cite{yan2023secure} can reduce but not eliminate this deviation and therefore do not satisfy the zero-KL requirement\cite{cachin1998information,hopper2002provably,qi2024provably}. A recent breakthrough, SyncPool~\cite{qi2024provably}, achieves provable security by replacing pruning with synchronization. However, its coarse-grained intra-pool synchronization imposes a high cost. It systematically discards the Shannon entropy of non-selected candidates, which substantially reduces embedding capacity. The loss grows as candidate pools become larger, and this severely limits practicality for short-form covert communication\cite{qi2024provably,li2024imperceptible}.

To address this trade-off between security and capacity, we introduce \method{}, a recursive disambiguation algorithm that remains within the synchronization paradigm while recovering capacity. Instead of synchronizing an entire ambiguous group, \method{} performs a minimal synchronized sample only over sequences that are indistinguishable to the receiver at the current visible step. The algorithm preserves all other discernible paths so that their entropy remains available for subsequent embedding. Our contributions are threefold:
\begin{enumerate}[leftmargin=*,label=(\arabic*),topsep=0pt,itemsep=0pt,parsep=0pt]
\item We design \method{}, a recursive disambiguation algorithm, and prove its computational zero-KL security.
\item We derive the theoretical capacity upper bound for zero-KL disambiguation and analyze the sources of gap between this bound and \method{}.
\item Using modern large language models, we have demonstrated that \method{} consistently approaches the capacity upper bound and substantially outperforms state-of-the-art baselines in embedding capacity.

\end{enumerate}

\section{Background}
\label{sec:background}

This section establishes the formal framework for our work by covering three key areas. First, we formalize the security standards that govern linguistic steganography. Second, we provide an in-depth account of the system architecture prevalent in modern ambiguity-aware methods. Finally, we review prior disambiguation architectures to situate our contribution within the current state of the art.

\subsection{Security Definitions in Steganography}
\label{subsec:security_definitions}

The foundational goal of linguistic steganography is to remain invisible to a watchful adversary. This adversarial setting is classically modeled by Simmons' \emph{Prisoners' Problem}, where two parties must communicate covertly under the surveillance of a warden~\cite{simmons1984prisoners,petitcolas2000information}. In this scenario, every transmission attempted by the prisoners is inspected. For each observed message, the warden must decide whether it is an ordinary letter or a coded note carrying concealed information. From a formal perspective, this is equivalent to a hypothesis test between the covertext distribution, $P_T$, and the stegotext distribution, $P_s$. A steganographic system is considered secure if it can render these two distributions statistically indistinguishable, and this requirement is formalized by two primary standards of security.

\paragraph{Information-Theoretic Security}
This standard represents the most stringent security guarantee, mandating that the stegotext and covertext distributions be mathematically identical. This is quantified by requiring the Kullback--Leibler (KL) divergence between them to be exactly zero~\cite{cachin1998information}:
\begin{equation}
    \label{eq:kl_divergence_prelim}
    D_{\mathrm{KL}}(P_s \parallel P_T) = \sum_{x \in \mathcal{X}} P_s(x) \log \frac{P_s(x)}{P_T(x)} = 0,
\end{equation}
where $\mathcal{X}$ is the set of all possible (terminal) messages, $P_T$ denotes the distribution of genuine covertext, and $P_s$ denotes the distribution of stegotext induced by the stegosystem under the same conditions. A system satisfying this condition achieves perfect security, as no adversary, regardless of their computational power, can gain an advantage in the decision problem.

\paragraph{Computational Security}
This standard offers a practical and rigorous guarantee for systems that employ cryptographic primitives. It defines security in the context of an adversary restricted to probabilistic polynomial-time (PPT) computations~\cite{hopper2002provably}. A system is considered computationally secure if the advantage any such adversary has in distinguishing between $P_s$ and $P_T$ is negligible, denoted as $\mathrm{negl}(\kappa)$ for a security parameter $\kappa$:
\begin{equation}
    \label{eq:comp_security_prelim}
    \left| \Pr[\mathcal{A}(s)=1] - \Pr[\mathcal{A}(t)=1] \right| < \mathrm{negl}(\kappa),
\end{equation}
where $s \sim P_s$, $t \sim P_T$, and $\mathcal{A}$ is any PPT adversary. Therefore, a primary objective for provably secure steganography in practice is to achieve a computational zero-KL guarantee, which ensures that the perfect security property of Eq.~\eqref{eq:kl_divergence_prelim} holds against any computationally bounded adversary.

\subsection{Secure Steganography in Autoregressive Models}
\label{subsec:steganography_in_llms}

The widespread adoption of large language models (LLMs) has revolutionized linguistic steganography. By generating fluent and contextually coherent text, these models provide an ideal source of covertext\cite{vaswani2017attention,brown2020language,touvron2023llama,li2025coas} that can mimic human writing.\cite{lin2024zero,huang2024od} The core embedding principle is a process known as message-driven sampling. In this process, an entropy encoder uses the model's output probability distribution and the secret bitstream at each step to select the next token, continuing autoregressively~\cite{ziegler2019neural,yang2018rnn,de2022perfectly}.

However, this steganographic paradigm is complicated by the practical necessity of tokenization. For transmission, the sender must \emph{detokenize} the generated token sequence into a human-readable string. The receiver, in turn, must \emph{retokenize} this string to recover the underlying message. Due to the nature of subword tokenizers, however, this retokenization is not guaranteed to reproduce the sender's original sequence. This potential for discrepancy is the core of the tokenization ambiguity problem~\cite{nozaki2022addressing}. Any such desynchronization corrupts the conditional probabilities for all subsequent steps, leading to a catastrophic decoding failure~\cite{qi2024provably}.

\begin{figure}[t]
    \centering
    \includegraphics[width=\columnwidth]{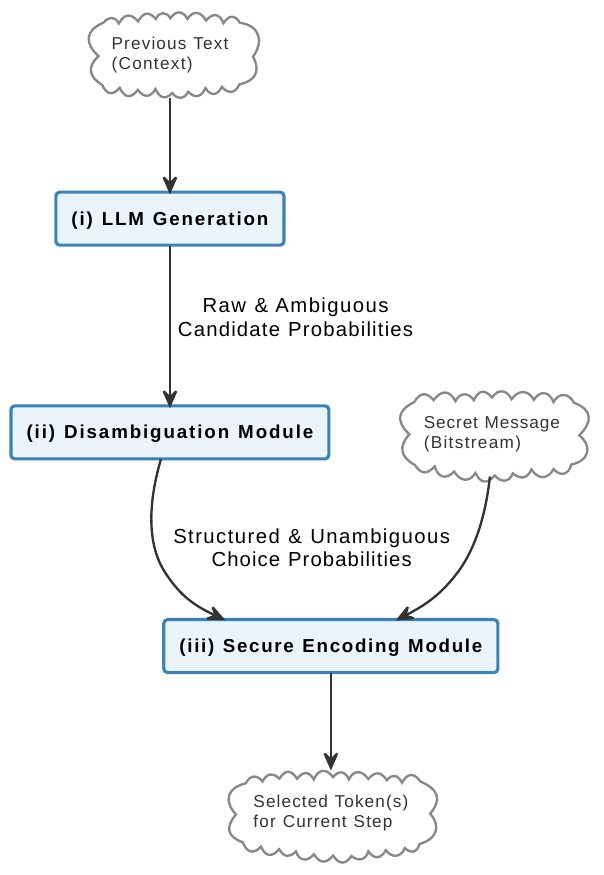}
    \caption{A schematic of the modular pipeline for a single generation step in an ambiguity-aware steganographic system, where the Disambiguation Module transforms the LLM's raw output into a structured set of choices for secure encoding.}
    \label{fig:steganographic_pipeline}
\end{figure}

To manage this challenge, modern ambiguity-aware systems adopt the modular pipeline depicted in Figure~\ref{fig:steganographic_pipeline}. A single generation step is decomposed into three distinct stages:
\begin{enumerate}[label=(\roman*), leftmargin=*]
    \item \textbf{LLM Generation.} The base language model takes the current context and produces a raw, and potentially ambiguous, probability distribution over its vocabulary.
    \item \textbf{Disambiguation Module.} A dedicated module transforms this raw distribution into a new, well-defined distribution over a set of unambiguous candidate choices, resolving all token-level conflicts before encoding.
    \item \textbf{Secure Encoding Module.} The entropy coder, guided by the secret message, samples a choice from the structured distribution provided by the disambiguation module to select the token(s) for the current step.
\end{enumerate}

For the end-to-end system to achieve zero-KL security, both the Secure Encoding and Disambiguation modules must independently be distribution-preserving. The Secure Encoding module must function as a perfect sampler, drawing from its input distribution without statistical bias. This challenge has been effectively addressed by state-of-the-art entropy coders such as iMEC and Discop~\cite{de2022perfectly,ding2023discop}. The Disambiguation Module faces the more recent challenge of structuring the choices so that the resulting probability space remains mathematically equivalent to the original. Methods such as SyncPool have shown that this is achievable~\cite{qi2024provably}, setting the stage for further architectural improvements.

\subsection{Prior Disambiguation Architectures}
\label{subsec:prior_architectures}

Existing designs for the disambiguation module can be categorized into two competing architectures.

\paragraph{Distribution-Altering Architectures}
This approach, first proposed for generative models by Nozaki and Murawaki~\cite{nozaki2022addressing}, resolves ambiguity by directly \emph{pruning} the candidate set. The mechanism removes any token that serves as a prefix for another candidate. Although this ensures decodability, it fundamentally alters the probability distribution. More sophisticated variants that employ a maximum-weight independent set (MWIS) to minimize the probability mass of the pruned tokens~\cite{yan2023secure} suffer from the same inherent flaw, since the act of deleting valid candidates creates a nonzero KL divergence and renders the entire architectural class insecure by definition.~\cite{zhang2021provably}.

\paragraph{Distribution-Preserving Architectures}
This architecture was introduced by SyncPool~\cite{qi2024provably}, the first module to achieve computational zero-KL security. The design involves a two-stage process. First, it groups all candidates with prefix relationships into ambiguity pools (e.g., \{\_B, \_BB, \_BBD\}). The encoder module embeds the payload by selecting one of these pools, which is an unambiguous choice. Second, to resolve the selection \emph{within} the chosen pool, the module employs a non-payload-bearing synchronized sampler. This sampler uses a cryptographically secure pseudorandom number generator (CSPRNG), seeded with a shared secret key, to choose a representative token according to its original probability distribution. While provably secure, this design consumes the Shannon entropy of the intra-pool selection for synchronization rather than for payload. This architectural inefficiency establishes the central technical challenge our work aims to solve.

This trade-off between provable security and embedding capacity defines the critical frontier for modern linguistic steganography. Addressing this architectural inefficiency is the primary motivation for our work. In the subsequent sections, we introduce \method{}, a disambiguation algorithm designed to retain the rigorous security guarantees of the synchronization-based paradigm while systematically recovering the capacity lost by existing methods.

\section{The \method{} Algorithm}
\label{sec:method}

\newcommand{\decode}{\operatorname{decode}}
\newcommand{\normalize}{\operatorname{normalize}}

The primary challenge for current provably secure steganography is resolving tokenization ambiguity. Existing secure methods, while avoiding statistical alterations, achieve this by aggressively discarding Shannon entropy, which leads to suboptimal embedding capacity. To overcome this limitation, we present \method{}, an algorithm designed to maximize capacity while upholding strict security guarantees. The core is a look-ahead resolution strategy that resolves only the necessary ambiguous cases, thereby preserving the information capacity of other distinguishable paths for subsequent embedding steps.

\begin{figure}[t]
  \centering
  \includegraphics[width=\columnwidth]{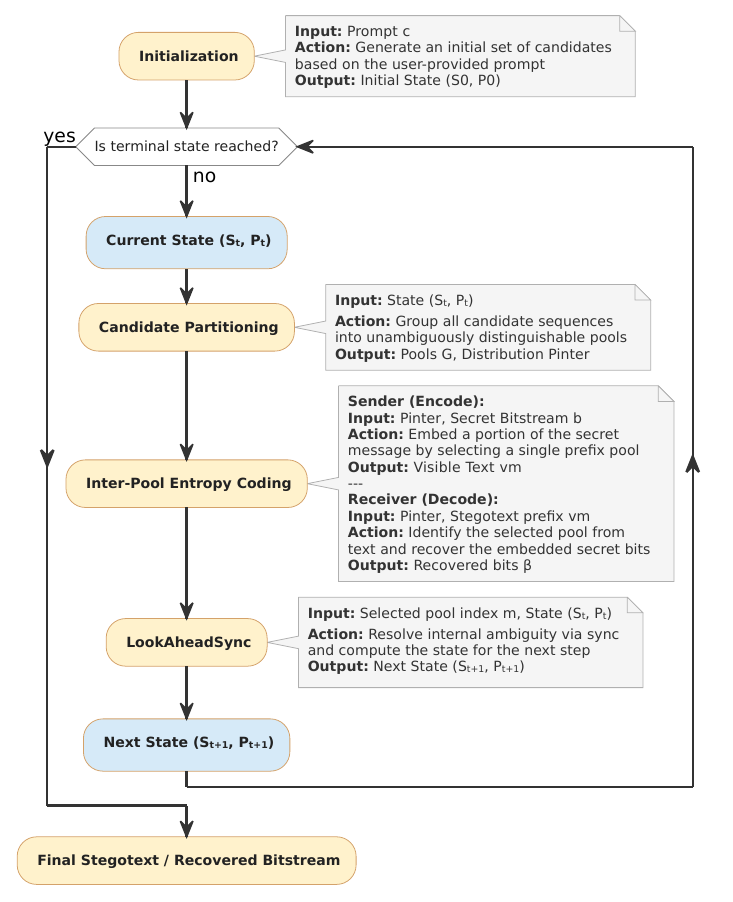}
  \caption{The iterative architecture of \method{}. The process begins with initialization and then enters a main loop that repeatedly executes three steps, namely partitioning candidates, embedding payload, and resolving ambiguity via a look-ahead mechanism to compute the next state. The loop continues until a terminal state is reached.}
  \label{fig:main_flowchart}
\end{figure}

This strategy is implemented within an iterative, state-driven architecture, as depicted in Figure~\ref{fig:main_flowchart}. The loop begins from the first set of  candidates produced by the base model under a prompt shared between the sender and the receiver. In each iteration, the algorithm embeds part of the secret bit and prepares the next round by executing three phases:

\begin{itemize}[leftmargin=*]
\item \textbf{Candidate Partitioning.} Group the model's candidate continuations by shared visible prefixes to obtain pools that are unambiguously distinguishable from one another, and aggregate within-pool probability mass to form an inter-group distribution.

\item \textbf{Inter-Group Entropy Coding.} Embed a portion of the secret by selecting exactly one pool according to the inter-group distribution at the sender, and recover the same bits by identifying the selected visible prefix at the receiver. This is the only payload-bearing step.

\item \textbf{Look-ahead Resolution.} Resolve any remaining ambiguity inside the selected pool by synchronizing only among truly indistinguishable sequences and performing a minimal look-ahead expansion, while preserving all already discernible continuations so their information remains available for future embedding.
\end{itemize}

We next detail these three operations and then present an end-to-end description of how they compose at the sender and the receiver.

\subsection{Candidate Partitioning}
\label{subsec:candidate_pooling}

The first operation in iteration $t$ partitions the global candidate set $S_t$ to prepare it for secure payload embedding. All candidate sequences are organized into disjoint groups according to their visible string prefixes. This partition ensures that each group is unambiguously distinguishable from the others, which is required by the subsequent encoding stage.

The mechanism proceeds in two steps. First, the algorithm applies the detokenization function $\varphi(\cdot)$ to each candidate sequence in $S_t$ to obtain its visible string representation. Second, it groups the original candidate sequences based on these strings. We form groups keyed by a visible string $v$ and assign to that group every candidate whose visible string starts with $v$ (if no other item shares the prefix, the group has one element). To enable a single linear pass, we sort candidates by their visible strings so that a shorter prefix appears before any longer string that starts with it; this guarantees that all items sharing the same prefix are contiguous.

Formally, the outcome is a partition of $S_t$ into $M$ disjoint groups, denoted by
\begin{equation}
    \mathcal{S}^{\text{inter}} \coloneqq \{\mathcal{S}^{\text{intra}}_0, \mathcal{S}^{\text{intra}}_1, \dots, \mathcal{S}^{\text{intra}}_{M-1}\}.
    \label{eq:s_inter_collection}
\end{equation}
With the candidates organized into these groups, we define the inter-group probability distribution $P^{\text{inter}}$, where each component $p^{\text{inter}}_m$ is the total probability mass of group $\mathcal{S}^{\text{intra}}_m$:
\begin{equation}
    p^{\text{inter}}_m \coloneqq \sum_{s \in \mathcal{S}^{\text{intra}}_m} P_t(s),
    \label{eq:p_inter_component}
\end{equation}
and collect these components, in the same order as the groups, into the vector
\begin{equation}
    P^{\text{inter}} \coloneqq (p^{\text{inter}}_0, p^{\text{inter}}_1, \dots, p^{\text{inter}}_{M-1}).
    \label{eq:p_inter_vector}
\end{equation}
Since $\{\mathcal{S}^{\text{intra}}_m\}_{m=0}^{M-1}$ is a partition of $S_t$ and $P_t$ is a probability distribution over $S_t$, we have $\sum_{m=0}^{M-1} p^{\text{inter}}_m = 1$. For decoding, we also record the ordered list of group keys (visible prefixes), denoted $V=[v_0,\dots,v_{M-1}]$.

\begin{algorithm}[H]
\footnotesize
\caption{\textsc{PartitionByPrefix}$(S_t, P_t)$}
\label{alg:partitioning}
\begin{algorithmic}[1]
\Require Candidate sequence set $S_t$; probability mapping $P_t$
\Ensure Collection of disjoint groups $\mathcal{S}^{\text{inter}}$; inter-group distribution $P^{\text{inter}}$; list of group keys $V$

\State Sort $S_t$ (and carry $P_t$ along) by the visible strings $\varphi(s)$ so that any shorter prefix appears before longer strings starting with it.
\State $\mathcal{S}^{\text{inter}} \leftarrow \emptyset$; \ $P^{\text{inter}} \leftarrow \emptyset$; \ $V \leftarrow \emptyset$
\If{$S_t$ is not empty}
    \State $v_{\text{prefix}} \leftarrow \varphi(S_t[0])$ \Comment{current group key}
    \State $\mathcal{S}^{\text{intra}}_{\text{current}} \leftarrow \{S_t[0]\}$
    \For{$i \leftarrow 1$ \textbf{to} $|S_t|-1$}
        \If{$\varphi(S_t[i])$ starts with $v_{\text{prefix}}$}
            \State Add $S_t[i]$ to $\mathcal{S}^{\text{intra}}_{\text{current}}$
        \Else
            \State Append $\mathcal{S}^{\text{intra}}_{\text{current}}$ to $\mathcal{S}^{\text{inter}}$
            \State Append $v_{\text{prefix}}$ to $V$
            \State $\mathcal{S}^{\text{intra}}_{\text{current}} \leftarrow \{S_t[i]\}$ \Comment{start new group}
            \State $v_{\text{prefix}} \leftarrow \varphi(S_t[i])$
        \EndIf
    \EndFor
    \State Append $\mathcal{S}^{\text{intra}}_{\text{current}}$ to $\mathcal{S}^{\text{inter}}$
    \State Append $v_{\text{prefix}}$ to $V$
\EndIf

\For{each group $\mathcal{S}^{\text{intra}}_m$ in $\mathcal{S}^{\text{inter}}$}
    \State $p_m \leftarrow \sum_{s \in \mathcal{S}^{\text{intra}}_m} P_t(s)$
    \State Append $p_m$ to $P^{\text{inter}}$
\EndFor

\State \Return $(\mathcal{S}^{\text{inter}}, P^{\text{inter}}, V)$
\end{algorithmic}
\end{algorithm}

\subsection{Inter-Group Entropy Coding}
\label{subsec:inter_entropy}

With the candidate space partitioned into $\mathcal{S}^{\text{inter}}$, the second operation embeds a segment of the secret payload by selecting one group. This is performed with a state-of-the-art entropy encoder, such as iMEC~\cite{de2022perfectly} or Discop~\cite{ding2023discop}, which we denote abstractly by $\mathcal{E}$. The sender applies $\mathcal{E}$ to the inter-group distribution $P^{\text{inter}}$ and the secret bitstream $b$, yielding the selected group's index $m$ and the consumed bits $\beta$. Symmetrically, the receiver identifies the index $m$ from the visible string and applies the inverse function $\mathcal{E}^{-1}$ to recover $\beta$:
\begin{align}
    (m, \beta) &\leftarrow \mathcal{E}\!\left(P^{\text{inter}},\, b\right), \label{eq:inter_encode} \\
    \beta &\coloneqq \mathcal{E}^{-1}\!\left(P^{\text{inter}},\, m\right). \label{eq:inter_decode}
\end{align}
Selecting group $m$ identifies the unique group key $v_m$ that is a prefix of the final stegotext $T$. The decoder finds $v_m$ as a prefix of the $T$, uses Eq.~\eqref{eq:inter_decode} to recover $\beta$.

While selecting $\mathcal{S}^{\text{intra}}_m$ completes the payload-bearing choice for this step, the ambiguity \emph{within} the chosen group remains unresolved. To prepare this group for the final resolution stage, its sub-distribution must be normalized so that subsequent choices are probabilistically sound. The inputs to the next stage are the chosen intra-group candidate set $\mathcal{S}^{\text{intra}}_m$ and its normalized distribution $P_m^{\text{intra}}$:
\begin{equation}
    P_m^{\text{intra}} \coloneqq \normalize\!\left(\{\,P_t(s) \mid s \in \mathcal{S}^{\text{intra}}_m\,\}\right),
    \label{eq:p_intra_norm}
\end{equation}
which is equivalently given by
\begin{equation}
    P_m^{\text{intra}}(s) \;=\; \frac{P_t(s)}{p_m^{\text{inter}}}
    \quad \text{for all } s \in \mathcal{S}^{\text{intra}}_m,
    \label{eq:p_intra_explicit}
\end{equation}
where $p_m^{\text{inter}}$ is defined in Eq.~\eqref{eq:p_inter_component}.

\begin{figure}[t]
  \centering
  \includegraphics[width=\columnwidth]{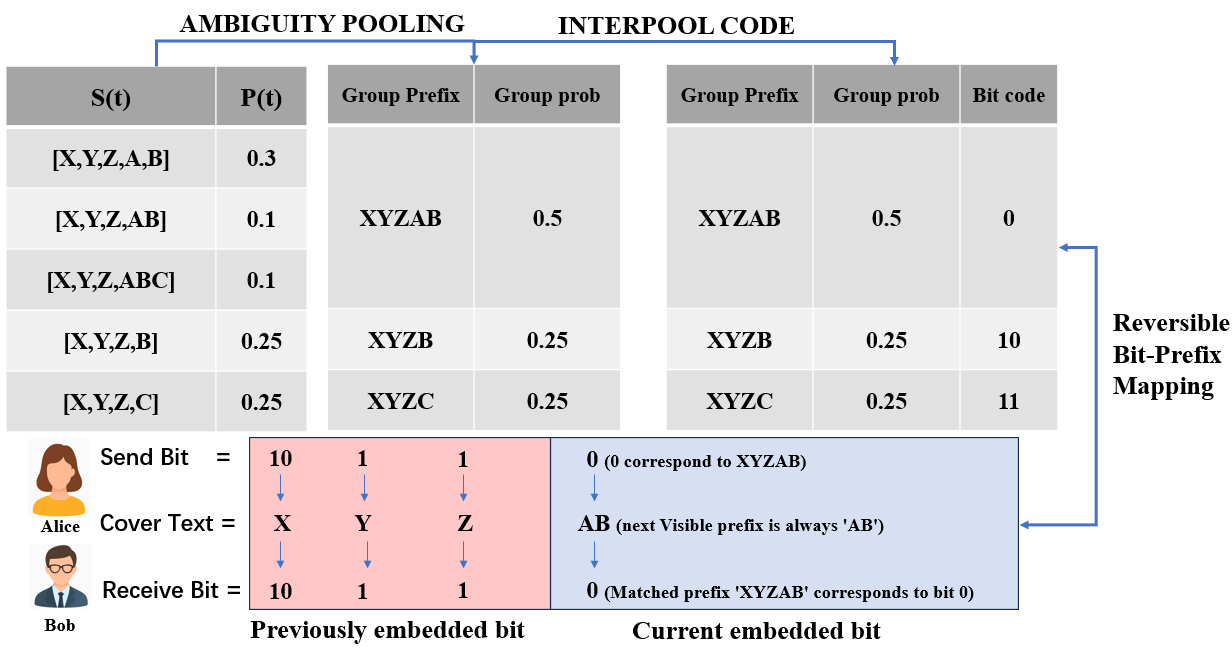}
  \caption{Inter-group encoding. Candidate token sequences are grouped by a common visible prefix and their probabilities are aggregated. An entropy encoder then maps a segment of the secret bitstream to a unique group according to the aggregated probabilities.}
  \label{fig:interpool-example}
\end{figure}

\begin{algorithm}[H]
\footnotesize
\caption{\textsc{LookAhead}}
\label{alg:lookahead}
\begin{algorithmic}[1]
\Require Intra-group set $\mathcal{S}^{\text{intra}}_m$; normalized distribution $P_m^{\text{intra}}$; shared key $\mathsf{K}$; language model \text{LLM}
\Ensure Next-round state $(S_{t+1}, P_{t+1})$ or a terminal sequence $s_{\text{end}}$

\State $v_m \gets \textsc{VisiblePrefix}(\mathcal{S}^{\text{intra}}_m)$
\State $\mathcal{S}_{\text{prefix}} \gets \{\, s \in \mathcal{S}^{\text{intra}}_m \mid \varphi(s)=v_m \,\}$
\State $\mathcal{S}_{\text{partial}} \gets \mathcal{S}^{\text{intra}}_m \setminus \mathcal{S}_{\text{prefix}}$

\State $P_{\text{prefix}} \gets \normalize\big(\{P_m^{\text{intra}}(s) : s \in \mathcal{S}_{\text{prefix}}\}\big)$
\State $s_{\text{sync}} \gets SyncSample(\mathcal{S}_{\text{prefix}}, P_{\text{prefix}}, \mathsf{K})$

\If{\textsc{IsEOS}($s_{\text{sync}}$) \textbf{and} $\mathcal{S}_{\text{partial}}=\varnothing$}
    \State \Return \textsc{End}($s_{\text{sync}}$) \Comment{terminate; wrapper for IsEnd/EndSeq}
\EndIf

\If{\textsc{IsEOS}($s_{\text{sync}}$)}
    \State $\mathcal{A}_{\text{next}} \gets \varnothing$; \ $P_{\text{next}} \gets \varnothing$
\Else
    \State $(\mathcal{A}_{\text{next}}, P_{\text{next}}) \gets \text{LLM}(\cdot \mid s_{\text{sync}})$
\EndIf

\State $p_{\text{sum}} \gets \sum_{s \in \mathcal{S}_{\text{prefix}}} P_m^{\text{intra}}(s)$
\State $S_{t+1} \gets \mathcal{S}_{\text{partial}}$
\ForAll{$x \in \mathcal{A}_{\text{next}}$}
    \State $S_{t+1} \gets S_{t+1} \cup \{\, s_{\text{sync}} \oplus x \,\}$
\EndFor

\State Define $P_{t+1}$ as a mapping as follows:
\ForAll{$s \in \mathcal{S}_{\text{partial}}$}
    \State $P_{t+1}(s) \gets P_m^{\text{intra}}(s)$
\EndFor
\ForAll{$x \in \mathcal{A}_{\text{next}}$}
    \State $P_{t+1}(s_{\text{sync}} \oplus x) \gets p_{\text{sum}} \cdot P_{\text{next}}(x)$
\EndFor

\State \Return $(S_{t+1}, P_{t+1})$
\end{algorithmic}
\end{algorithm}

\subsection{Look-ahead Resolution}
\label{subsec:lookahead}

While the selection of a group in the previous step is an unambiguous, payload-bearing choice, the ambiguity \emph{within} the chosen group $\mathcal{S}^{\text{intra}}_m$ remains unresolved. All candidate sequences in this set share the same visible prefix, which makes them indistinguishable to the receiver at the current step. This indistinguishability blocks further secure embedding and must be resolved before the next round. The objective in this operation is to perform a \emph{distribution-preserving transformation} of the candidate space, converting the ambiguous input state $(\mathcal{S}^{\text{intra}}_m, P_m^{\text{intra}})$ into a new, unambiguous state $(S_{t+1}, P_{t+1})$ on which the next payload-bearing decision can safely proceed.

We address the ambiguity at its source. Let $v_m$ denote the common visible prefix of all sequences in $\mathcal{S}^{\text{intra}}_m$. The indistinguishability arises precisely from those sequences that decode \emph{exactly} to $v_m$. The look-ahead mechanism resolves this by generating new, distinct continuations from that source, thereby separating paths that were previously identical at the visible level while preserving probability mass.

Formally, the algorithm first partitions $\mathcal{S}^{\text{intra}}_m$ into two functionally distinct subsets. The \emph{Prefix Set} collects the exact-prefix items that decode to $v_m$ and will serve as the input context for look-ahead; the \emph{Partial Set} contains longer continuations that are already distinguishable and will be preserved to the next round:
\begin{align}
    \mathcal{S}_{\text{prefix}}
      &:= \bigl\{\,s \in \mathcal{S}^{\text{intra}}_{m} \;\big|\; \varphi(s) = v_{m}\bigr\}, \label{eq:prefix_set}\\
    \mathcal{S}_{\text{partial}}
      &:= \mathcal{S}^{\text{intra}}_{m} \setminus \mathcal{S}_{\text{prefix}}. \label{eq:partial_set}
\end{align}

To remain faithful to the original distribution while avoiding a combinatorial explosion, we select a \emph{single} synchronized representative from the Prefix Set using a non-payload-bearing sampler. Concretely, we normalize the intra-set probabilities to obtain a valid sampling distribution $P_{\text{prefix}}$ and then draw $s_{\text{sync}}$ with a CSPRNG-seeded synchronized procedure:
\begin{equation}
    P_{\text{prefix}} := \normalize\!\left(\{\,P_m^{\text{intra}}(s) \mid s \in \mathcal{S}_{\text{prefix}}\,\}\right),
    \label{eq:normalize_prefix}
\end{equation}
\begin{equation}
    s_{\text{sync}} := SyncSample\!\left(\mathcal{S}_{\text{prefix}},\, P_{\text{prefix}},\, K\right).
    \label{eq:sync_sample}
\end{equation}
where SyncSample$(\mathcal{S}, P, K)$ denotes a shared, distribution-preserving sampler driven by a CSPRNG initialized with the shared key $K$.

With $s_{\text{sync}}$ fixed, the algorithm performs a single deterministic forward pass of the base model conditioned on this representative. This yields a set of next-token candidates and their conditional probabilities:
\begin{equation}
    (\mathcal{A}_{\text{next}},\, P_{\text{next}}) := \text{LLM}(\cdot \mid s_{\text{sync}}),
    \label{eq:llm_forward_pass}
\end{equation}
where $P_{\text{next}}$ is a probability distribution over $x \in \mathcal{A}_{\text{next}}$ and satisfies $\sum_{x \in \mathcal{A}_{\text{next}}} P_{\text{next}}(x)=1$.

We then merge the preserved partials with the freshly expanded continuations to construct the next state. Let
\begin{equation}
    p_{\text{sum}} := \sum_{s \in \mathcal{S}_{\text{prefix}}} P_m^{\text{intra}}(s),
    \label{eq:prob_sum}
\end{equation}
be the total probability mass of the Prefix Set under $P_m^{\text{intra}}$. The next candidate set concatenates each $x \in \mathcal{A}_{\text{next}}$ to the synchronized representative and unions the result with the Partial Set:
\begin{equation}
    S_{t+1} := \mathcal{S}_{\text{partial}} \;\cup\; \bigl\{\,s_{\text{sync}} \oplus x \;\big|\; x \in \mathcal{A}_{\text{next}} \bigr\},
    \label{eq:s_update}
\end{equation}
where $\oplus$ denotes sequence concatenation. The corresponding probabilities follow the law of total probability. We carry over masses for preserved partials and reallocate the entire Prefix-Set mass onto the new continuations by scaling $P_{\text{next}}$:
\begin{equation}
\begin{aligned}
    P_{t+1}(s) :=
    \begin{cases}
      P_m^{\text{intra}}(s), & \text{if } s \in \mathcal{S}_{\text{partial}},\\[4pt]
      p_{\text{sum}} \cdot P_{\text{next}}(x), & \text{if } s = s_{\text{sync}} \oplus x,\; x \in \mathcal{A}_{\text{next}}.
    \end{cases}
\end{aligned}
\label{eq:p_update}
\end{equation}
Since $\sum_{s \in \mathcal{S}_{\text{partial}}} P_m^{\text{intra}}(s) = 1 - p_{\text{sum}}$ and $\sum_{x \in \mathcal{A}_{\text{next}}} P_{\text{next}}(x)=1$, it follows that $\sum_{s \in S_{t+1}} P_{t+1}(s) = (1 - p_{\text{sum}}) + p_{\text{sum}} = 1$.

The resulting state $(S_{t+1}, P_{t+1})$ is thus a new, probabilistically sound candidate space in which the immediate ambiguity tied to $v_m$ has been eliminated without altering the overall distribution. We treat EOS as a visible string that cannot be extended; termination occurs if and only if the synchronized choice is EOS and the partial set is empty. All forward passes are deterministic; the only randomness arises from the synchronized sampler and the entropy coder, which are shared between sender and receiver.

\begin{figure*}[t]
  \centering
  \includegraphics[width=\linewidth]{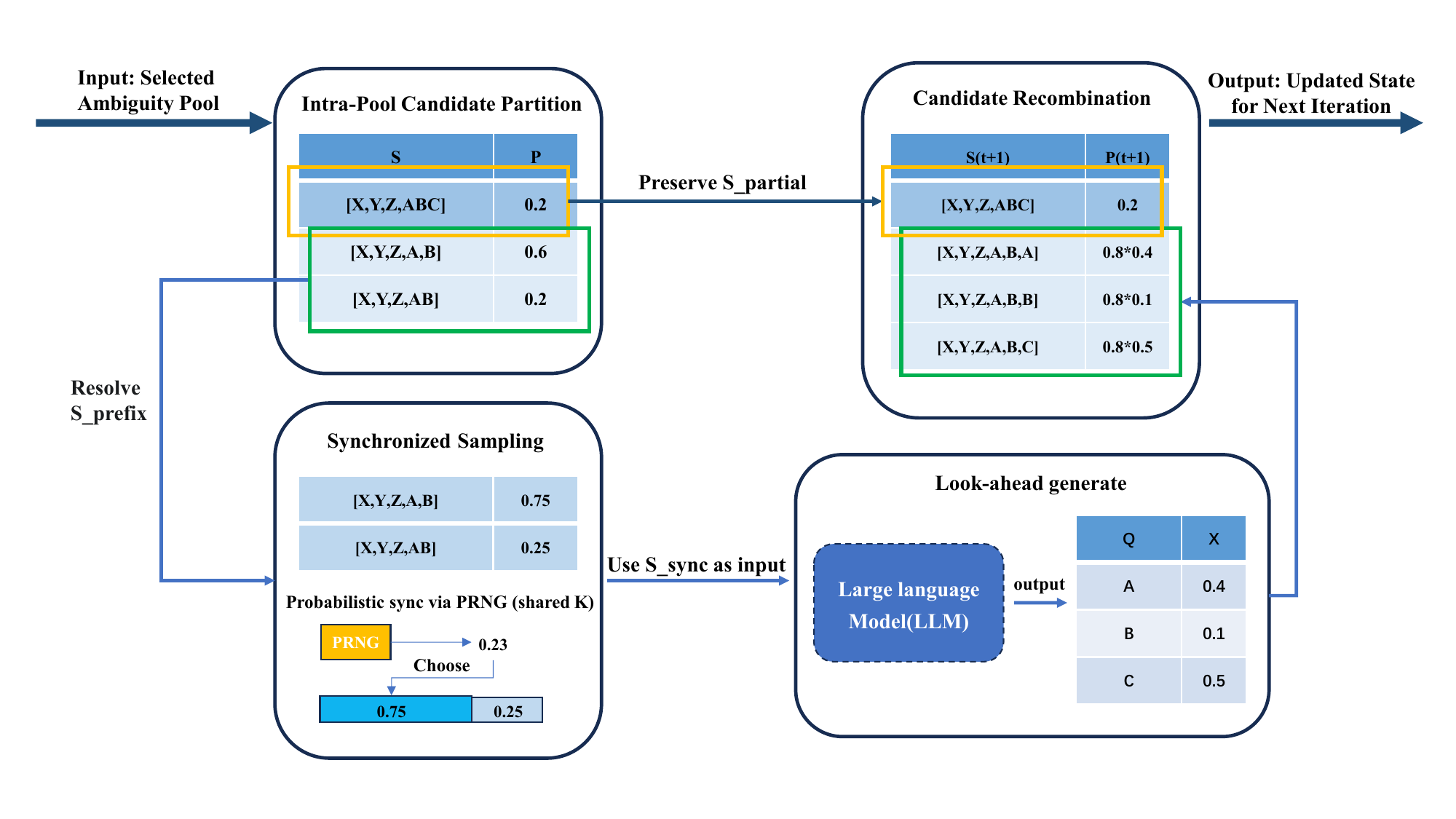}
  \caption{Look-ahead disambiguation. The selected intra-group candidates are partitioned into a prefix set and a partial set. A synchronized sampler selects a representative $s_{\text{sync}}$ from the prefix set, which is then expanded via an LLM call. The new candidate set for the next round is obtained by combining the preserved partial set with the new expansions, thereby reallocating probability mass to new continuations.}
  \label{fig:lookahead-example}
\end{figure*}

\subsection{End-to-End Steganographic Pipeline}
\label{sec:full_pipeline}

\begin{algorithm}[H]
\footnotesize
\caption{\textsc{EmbedLoop}}
\label{alg:embed_loop}
\begin{algorithmic}[1]
\Require Prompt $\mathbf{c}$, secret bitstream $b$
\Ensure Stegotext
\State $(S_t, P_t) \gets \text{LLM}(\mathbf{c})$ \Comment{initialization}
\While{true}
    \State $(\mathcal{S}^{\text{inter}}, P^{\text{inter}}, V) \gets \textsc{PartitionByPrefix}(S_t, P_t)$
    \State $(m, \beta) \gets \textsc{InterGroupEncode}(P^{\text{inter}}, b)$
    \State $b \gets \textsc{Consume}(b, \beta)$

    \State $\mathcal{S}^{\text{intra}}_m \gets \mathcal{S}^{\text{inter}}[m]$
    \State $P_m^{\text{intra}} \gets \normalize(\{P_t(s) \mid s \in \mathcal{S}^{\text{intra}}_m\})$

    \State $\rho \gets \textsc{LookAhead}(\mathcal{S}^{\text{intra}}_m, P_m^{\text{intra}}, \mathsf{K}, \text{LLM})$

    \If{\textsc{IsEnd}($\rho$)}
        \State \Return $\varphi(\textsc{EndSeq}(\rho))$
    \Else
        \State $(S_{t+1}, P_{t+1}) \gets \rho$
        \State $S_t \gets S_{t+1}$; \ $P_t \gets P_{t+1}$
    \EndIf
\EndWhile
\end{algorithmic}
\end{algorithm}

Having detailed the constituent modules of a single generation step, we now present the complete, end-to-end steganographic pipeline. These modules are integrated into the iterative loop described above and are specified for the sender in Algorithm~\ref{alg:embed_loop} and for the receiver in Algorithm~\ref{alg:decode_loop}. Let $T$ denote the final stegotext observed by the decoder; write $\epsilon$ for the empty bitstring and use $\Vert$ for bit concatenation. We also use $\rho$ to denote either a terminal marker or the next-state pair $(S_{t+1}, P_{t+1})$. 

\begin{algorithm}[H]
\footnotesize
\caption{\textsc{DecodeLoop}}
\label{alg:decode_loop}
\begin{algorithmic}[1]
\Require Prompt $\mathbf{c}$, stegotext $T$ \Comment{$T$ is immutable}
\Ensure Recovered bitstream $\hat{b}$
\State $(S_t, P_t) \gets \text{LLM}(\mathbf{c})$ \Comment{initialization}
\State $\hat{b} \gets \epsilon$ \Comment{empty bitstream}
\While{true}
    \State $(\mathcal{S}^{\text{inter}}, P^{\text{inter}}, V) \gets \textsc{PartitionByPrefix}(S_t, P_t)$
    \State $m \gets \textsc{MatchPrefixIndex}(V, T)$ \Comment{find the unique $v_m \in V$ that prefixes the same $T$}
    \State $\beta \gets \textsc{InterGroupDecode}(P^{\text{inter}}, m)$
    \State $\hat{b} \gets \hat{b} \Vert \beta$

    \State $\mathcal{S}^{\text{intra}}_m \gets \mathcal{S}^{\text{inter}}[m]$
    \State $P_m^{\text{intra}} \gets \normalize(\{P_t(s) \mid s \in \mathcal{S}^{\text{intra}}_m\})$

    \State $\rho \gets \textsc{LookAhead}(\mathcal{S}^{\text{intra}}_m, P_m^{\text{intra}}, \mathsf{K}, \text{LLM})$

    \If{\textsc{IsEnd}($\rho$)}
        \State \Return $\hat{b}$
    \Else
        \State $(S_{t+1}, P_{t+1}) \gets \rho$
        \State $S_t \gets S_{t+1}$; \ $P_t \gets P_{t+1}$
    \EndIf
\EndWhile
\end{algorithmic}
\end{algorithm}

The structural identity of the embedding and decoding loops is the foundation of the system's reliability. The only functional difference lies in their interaction with the entropy coder. The \textsc{EmbedLoop} calls an encoder to consume bits, whereas the \textsc{DecodeLoop} calls a decoder to recover bits from the visible string. All other state transitions, including candidate partitioning and the synchronized sampling inside \textsc{LookAhead}, are deterministic functions of the public model state and the shared secret key $\mathsf{K}$. This ensures that the receiver follows the same execution path as the sender and that the recovered bitstream $\hat{b}$ is identical to the original bitstream $b$.

In summary, \method{} operates as a recursive, state-driven process that separates payload embedding from ambiguity resolution. Payload is embedded only during the unambiguous inter-group selection, whereas the subsequent intra-group resolution is handled by the look-ahead mechanism. This mechanism preserves distinguishable continuations and uses a single, non-payload-bearing synchronized sample to resolve only the core source of ambiguity. The architecture is designed to minimize the loss of Shannon entropy compared with prior work and thereby substantially improves embedding capacity.

\section{Security and Capacity Analysis}
\label{sec:analysis}
Having detailed the mechanics of the \method{} algorithm, we now present the theoretical analysis that supports its design and guarantees.
\begin{itemize}[leftmargin=*]
  \item \textbf{Security Analysis.} We prove that the algorithm upholds the standard of computational zero-KL security, ensuring its output is statistically indistinguishable from genuine text.
  \item \textbf{Capacity Analysis.} We derive the embedding-capacity upper bound for zero-KL disambiguation and quantify the gap between this bound and \method{}.
\end{itemize}

To ensure consistency and rigor in subsequent analyses, we first formalize the notation employed in this section:

Let $\mathcal{V}$ denote the set of all terminal visible strings.

Let $\varphi(\cdot)$ denote the detokenization function that maps a token sequence to a visible string.

All text generation is conditioned on an initial prompt $\mathbf{c}$; equivalently, all probabilities (e.g., $P_\theta(\cdot \mid \mathbf{c})$) are taken conditional on $\mathbf{c}$ throughout.

\subsection{Computational Zero-KL Security}
\label{subsec:proof-security}

The central security guarantee of \method{} is its computational security, which we formally prove in this section. The proof analyzes the algorithm's end-to-end generative process. We demonstrate that the probability of any given terminal sequence $s_{\text{end}}$ being generated by the algorithm is computationally indistinguishable from its true probability under the base language model, $P_\theta(s_{\text{end}} \mid \mathbf{c})$. The validity of this proof is contingent on the standard cryptographic assumption that the output of a cryptographically secure pseudorandom number generator (CSPRNG) is computationally indistinguishable from a truly random sequence to any probabilistic polynomial-time (PPT) adversary.\cite{haastad1999pseudorandom}

The proof proceeds by induction on an invariant that holds throughout the algorithm's state-driven execution. At each step $t$, the algorithm maintains a state $(S_t, P_t)$, where $S_t$ is the set of candidate sequences and $P_t$ is a probability mapping over $S_t$. As this state is determined by the history of probabilistic choices, both $S_t$ and $P_t$ are random variables. For clarity in this proof, we use the functional notation $P_t(s)$ to denote the weight corresponding to a sequence $s \in S_t$. A single instance of the state will therefore not align with the base model's distribution. To prove the security of the overall generative process, our analysis instead focuses on its behavior in expectation.

The invariant is defined over the expected values of these weights. It asserts that for any candidate sequence $s \in S_t$, the expectation of its corresponding weight $P_t(s)$ is computationally indistinguishable from the sequence's true probability under the base model $P_\theta$:
\begin{equation}
  \mathbb{E}\bigl[P_t(s)\bigr] \approx_c P_\theta(s \mid \mathbf{c}).
  \label{eq:security-inv}
\end{equation}
Throughout this section, $\approx_c$ is interpreted in the sense of computational indistinguishability defined in Eq.~\eqref{eq:comp_security_prelim}. When applied to real-valued quantities such as expectations, it indicates that the absolute deviation is bounded by a negligible function $\mathrm{negl}(\kappa)$. The expectation $\mathbb{E}$ is taken over the space of all possible random histories up to step $t$, which are determined by the probabilistic choices made by the entropy encoder and the synchronized sampler in all preceding steps.

The invariant holds for the base case ($t=0$), as the initial state $(S_0, P_0)$ is generated directly by the language model via $\text{LLM}(\mathbf{c})$. In this case, there is no random history, and for any $s \in S_0$, the equality $\mathbb{E}[P_0(s)] = P_0(s) = P_\theta(s \mid \mathbf{c})$ holds exactly.

For the inductive step, we assume the invariant holds for step $t$ and demonstrate that it also holds for step $t+1$. Any sequence $s' \in S_{t+1}$ is formed in one of two ways:
\begin{enumerate}[label=(\roman*)]
    \item Preservation: If $s'$ is preserved from a partial set, its weight $P_{t+1}(s')$ is non-zero only if its containing group $m$ is selected. By applying the law of total expectation over all possible group selections and substituting the definitions for $p^{\text{inter}}_m$ and $P_m^{\text{intra}}(s')$, the expression for the expected weight simplifies directly to its value from the previous step:
    \begin{equation}
        \mathbb{E}[P_{t+1}(s')] = \mathbb{E}[P_t(s')].
    \end{equation}
    By the inductive hypothesis, $\mathbb{E}[P_t(s')] \approx_c P_\theta(s' \mid \mathbf{c})$, thus preserving the invariant.

    \item Generation: If $s'$ is newly generated as an expansion $s' \coloneqq s_{\text{sync}} \oplus x$, its weight is defined as $P_{t+1}(s') \coloneqq p_{\text{sum}} \cdot P_\theta(x \mid s_{\text{sync}})$. The expectation is taken over the choice of group $m$ and the choice of $s_{\text{sync}}$. The selection of $s_{\text{sync}}$ via SyncSample is driven by the CSPRNG, making its expected behavior computationally indistinguishable from a true random sample over the prefix set $\mathcal{S}_{\text{prefix},m}$. This allows us to express the expectation as a sum over the members of that prefix set. By the linearity of expectation, we can move the summation outside:
    \begin{equation}
        \mathbb{E}[P_{t+1}(s')] \approx_c \sum_{s_p \in \mathcal{S}_{\text{prefix},m}} \mathbb{E}\left[ P_t(s_p) \cdot P_\theta(x \mid s_p) \right].
    \end{equation}
    Since $P_\theta(x \mid s_p)$ is a constant with respect to the outer expectation, it can be factored out:
    \begin{equation}
        \mathbb{E}[P_{t+1}(s')] \approx_c \sum_{s_p \in \mathcal{S}_{\text{prefix},m}} \mathbb{E}[P_t(s_p)] \cdot P_\theta(x \mid s_p).
    \end{equation}
    By the inductive hypothesis, $\mathbb{E}[P_t(s_p)] \approx_c P_\theta(s_p \mid \mathbf{c})$. Substituting this in and applying the chain rule of probability, we get:
    \begin{align}
        \mathbb{E}[P_{t+1}(s')] &\approx_c \sum_{s_p \in \mathcal{S}_{\text{prefix},m}} P_\theta(s_p \mid \mathbf{c}) \cdot P_\theta(x \mid s_p) \\
        &= P_\theta(s' \mid \mathbf{c}).
    \end{align}
    The invariant is therefore maintained for newly generated sequences.
\end{enumerate}
Since the invariant holds for both cases, it holds for all sequences in $S_{t+1}$. By induction, the probability of generating any terminal sequence $s_{\text{end}}$ is computationally indistinguishable from its true model probability. Therefore, the overall output distribution of \method{} is computationally secure.

\subsection{Theoretical Capacity Upper Bound}
\label{subsec:capacity-bound}

A rigorous analysis of the embedding capacity for provably secure linguistic steganography necessitates an understanding of its theoretical limits. This section derives the capacity upper bound for any steganographic system that resolves tokenization ambiguity while adhering to the zero-KL security standard. The bound provides a benchmark against which such methods, including our own, can be measured.

Perfect security dictates that the distribution of generated terminal sequences, $P_{\mathrm{embed}}$, must be identical (or computationally indistinguishable) to the distribution induced by the original language model. A critical implication is that the distribution over the final, visible strings must also be preserved. Recall that $\mathcal{V}$ denotes the set of all possible terminal visible strings, and define the probability of generating a specific visible string $v \in \mathcal{V}$ as:
\begin{equation}
    P_{V_{\text{end}}}(v) \coloneqq \sum_{s_{\text{end}} :\, \varphi(s_{\text{end}})=v} P_\theta(s_{\text{end}} \mid \mathbf{c}).
\end{equation}
The zero-KL security constraint requires that the distribution over visible string produced by the steganographic system, $P_{\mathrm{embed}}(v)$, equals $P_{V_{\text{end}}}(v)$:
\begin{align}
    P_{\mathrm{embed}}(v) &= \sum_{s_{\text{end}} :\, \varphi(s_{\text{end}})=v} P_{\mathrm{embed}}(s_{\text{end}}) \nonumber \\
    &= \sum_{s_{\text{end}} :\, \varphi(s_{\text{end}})=v} P_\theta(s_{\text{end}} \mid \mathbf{c}) = P_{V_{\text{end}}}(v).
    \label{eq:visible_dist_preserve}
\end{align}
Any secure system is thus a perfect sampler from the model's distribution over observable, visible string. To quantify the embedding performance of such a system, we employ two standard metrics. Let $B(s_{\text{end}})$ be the total number of bits embedded during the generation of a terminal sequence $s_{\text{end}}$. The expected number of bits per sequence (BPS) and bits per token (BPT) are:
\begin{align}
    \mathrm{BPS} &\coloneqq \mathbb{E}_{s_{\text{end}} \sim P_\theta} \bigl[ B(s_{\text{end}}) \bigr], \label{eq:bps} \\
    \mathrm{BPT} &\coloneqq \frac{\mathrm{BPS}}{\mathbb{E}_{s_{\text{end}} \sim P_\theta}[|s_{\text{end}}|]}. \label{eq:bpt}
\end{align}
where $|s_{\text{end}}|$ denotes the number of tokens in the terminal sequence $s_{\text{end}}$.

To analyze the capacity bounds of the disambiguation strategy itself, independent of a specific coder's implementation, we employ the theoretical construct of an ideal entropy encoder.\cite{rissanen1976generalized} Such a coder is perfectly efficient, meaning the expected number of bits it embeds is exactly equal to the Shannon entropy of the target distribution.\cite{shannon1948mathematical} Since a secure system must sample from the distribution of visible string $P_{V_{\text{end}}}$, Shannon's source coding theorem dictates that the maximum average number of bits that can be encoded is bounded by the entropy of this distribution,
\[
  H(V_{\text{end}}) \;=\; -\sum_{v \in \mathcal{V}} P_{V_{\text{end}}}(v) \log_2 P_{V_{\text{end}}}(v),
\]
see, e.g., \cite{cover1999elements}. This yields the upper bounds:
\begin{align}
    \mathrm{BPS} &\le H(V_{\text{end}}), \label{eq:bps_bound} \\
    \mathrm{BPT} &\le \frac{H(V_{\text{end}})}{\mathbb{E}_{s_{\text{end}} \sim P_\theta}[|s_{\text{end}}|]}. \label{eq:capacity_bound}
\end{align}
These bounds represent the maximum embedding rates achievable by any disambiguation algorithm that satisfies the zero-KL security criterion.

\subsection{Analysis of \method{}'s Capacity}
\label{subsec:optimality-gap}

Having established the theoretical capacity upper bound, we now analyze the performance of the \method{} algorithm against this benchmark. The payload capacity of our method is derived exclusively from the entropy of the inter-group selection process at each step $t$. However, a gap exists between the achieved capacity and this theoretical optimum. This discrepancy is an inherent consequence of the non-payload-bearing synchronization mechanism required to resolve intra-group ambiguity.

This capacity gap materializes during the SyncSample step. Although the member sequences of the prefix set $\mathcal{S}_{\text{prefix}}$ are mapped to an identical visible string at the current step, they function as distinct contexts for subsequent generation. The statistical divergence in their future outcomes could, in principle, be leveraged to encode information. To exploit this, a globally optimal algorithm would need to recursively expand every sequence within $\mathcal{S}_{\text{prefix}}$ to explore all possible futures, leading to a combinatorial explosion of candidate paths. In contrast, our algorithm deliberately forgoes this exhaustive expansion, instead using SyncSample to select a single representative path for feasibility.

To analyze this forgone capacity, we define the synchronization loss as the difference between the maximum theoretical capacity available at a given step and the actual capacity achieved by our algorithm. At a step where a group with visible prefix $v_m$ is chosen, the maximum future capacity is the entropy of the subsequent visible outcomes $V_{\text{end}}$ conditioned only on this public information, namely $H(V_{\text{end}} \mid v_m)$. Our algorithm's actual capacity, however, is further conditioned on the specific choice made by SyncSample. We model this choice as a random variable $S_{\text{prefix}}$ over the prefix set $\mathcal{S}_{\text{prefix}}$. Thus, the actual capacity is the expected conditional entropy, averaged over all possible synchronized choices:
\begin{equation}
    H_{\text{loss}}^{\text{sync}} \coloneqq H(V_{\text{end}} \mid v_m) - H(V_{\text{end}} \mid S_{\text{prefix}}, v_m).
\end{equation}

This definition has a direct interpretation in information theory. Based on the chain rule for conditional entropy, this difference is the conditional mutual information between the synchronized choice $S_{\text{prefix}}$ and the future visible outcomes $V_{\text{end}}$, given the common prefix $v_m$\cite{mcgill1954multivariate}:
\begin{equation}
\begin{aligned}
H\!\left(V_{\text{end}}\mid S_{\text{prefix}}, v_m\right)
&\coloneqq \sum_{s_p \in \mathcal{S}_{\text{prefix}}} P(s_p \mid v_m)\, \\
&\quad \times H\!\left(V_{\text{end}} \mid S_{\text{prefix}}=s_p,\, v_m\right).
\end{aligned}
\end{equation}
The synchronization loss at this step is therefore
\begin{equation}
    H_{\text{loss}}^{\text{sync}} \equiv I(S_{\text{prefix}} ; V_{\text{end}} \mid v_m).
    \label{eq:sync_loss}
\end{equation}
This equivalence provides a formal tool to reason about the capacity gap by revealing that the loss is the amount of information about the final outcome revealed by the non-payload-bearing synchronization process itself. The total gap between our algorithm's performance and the theoretical bound is the sum of these expected losses over the entire generation process.

An analysis of the synchronization loss in Eq.~\eqref{eq:sync_loss} shows that its magnitude is determined by the statistical divergence among the future text distributions conditioned on different sequences within the ambiguous prefix set $\mathcal{S}_{\text{prefix}}$. A significant loss arises only when the distributions that follow semantically equivalent but tokenization distinct sequences, such as [\_mis, \_trust] versus [\_mistrust], differ substantially. This observation motivates the Linguistic Smoothness Hypothesis, which states that a model’s semantic understanding yields statistically similar conditional distributions for such semantically equivalent sequences, so the divergence is small and the resulting conditional mutual information in Eq.~\eqref{eq:sync_loss} is negligible. We provide empirical evidence in Section~\ref{sec:exp}.

\section{Experiments}
\label{sec:exp}

Following the theoretical analysis of \method{}'s security and capacity limits in Section~\ref{sec:analysis}, we now turn to its empirical evaluation. This section presents a set of experiments to validate our central theoretical claims. The evaluation is designed to confirm that \method{} maintains zero-KL security while achieving a high embedding capacity. We assess our method against key baselines on both English and Chinese benchmarks. The experiments also provide direct empirical support for the Linguistic Smoothness Hypothesis, which, as we have argued, underpins our algorithm's high efficiency.
\subsection{Experimental Setup}
\label{subsec:setup}

To empirically evaluate our proposed method, we conduct steganography experiments on both English and Chinese. These languages were chosen to ensure a robust assessment across different linguistic characteristics.

For English tasks, we employ the \textbf{Llama3-8b} model, and for Chinese, we utilize the \textbf{Qwen2.5-7b} model.\cite{grattafiori2024llama,hui2024qwen2} Both models are large language models based on Transformer architectures\cite{vaswani2017attention}, and their tokenizers are implemented based on subwords\cite{sennrich2015neural,kudo2018sentencepiece}, which is critical for the tokenization ambiguity problem addressed in this work. To intuitively assess how performance is affected by the number of elements in the candidate pool, we exclusively use top-$k$ sampling\cite{holtzman2019curious} to constrain the size of the initial candidate pool and its probability distribution. The temperature is consistently fixed at 1.0, as it does not directly affect the number of tokens. In our experiments, we set the truncation parameter $k \in \{16, 32, 128\}$.

For evaluation data, we sample 50 unique reviews from the IMDB movie review dataset for English tasks and 50 unique short-comments from the Douban dataset for Chinese tasks. Each sampled review or comment is truncated to serve as an initial context. This context is then provided as a prompt to the corresponding language model, which is tasked with generating a continuation. For each of these 50 unique prompts and for every top-$k$ setting, we generate 100 steganographic messages. Each message is embedded using a new, independently generated random bitstream as the secret payload, ensuring that our results are averaged over a diverse set of contexts and messages.

We benchmark \method{} against a carefully selected set of baselines. These include \textbf{MWIS Pruning}~\cite{yan2023secure}, a state-of-the-art distribution-altering method, and \textbf{SyncPool}~\cite{qi2024provably}, which represents the current state of the art in provably secure disambiguation. Additionally, we include results from applying the entropy encoder directly to the raw LLM output without any disambiguation, serving as a practical upper bound for capacity. To demonstrate broad compatibility, all methods are implemented on top of two distinct back-end entropy encoders: iMEC~\cite{de2022perfectly} and Discop~\cite{ding2023discop}. All experiments are carried out on NVIDIA RTX 4090 GPUs.

In a separate experiment designed to empirically test our Linguistic Smoothness Hypothesis, we analyze the statistical divergence within real-world ambiguity scenarios. For each model, we collect 1,000 instances where an ambiguity pool, or prefix set $\mathcal{S}_{\text{prefix}}$, contains at least two distinct token sequences that decode to the same visible string. For each instance, we then compute the generalized Jensen-Shannon Divergence (JSD)\cite{lin2002divergence} among the next-token probability distributions conditioned on each of these ambiguous sequences. This process yields an empirical distribution of JSD scores, which allows us to quantify the practical impact of the Synchronization Loss.

\subsection{Evaluation Metrics}
\label{subsec:metrics}

We evaluate all methods based on four key metrics, each designed to assess a critical aspect of the steganographic system's performance.

\begin{itemize}[leftmargin=*]
    \item \textbf{Embedding Capacity (BPT ↑)}: The primary metric for steganographic efficiency is bits Per Token (BPT). It is calculated by dividing the total number of embedded secret bits by the total number of tokens in the generated stegotext. A higher BPT value signifies a more efficient utilization of the language channel for data hiding.

    \item \textbf{Statistical Security (KL ↓)}: We quantify the statistical security by measuring the Kullback-Leibler (KL) Divergence between the probability distribution used for steganographic sampling and the original distribution from the base model. A KL divergence of zero indicates that the stego and cover distributions are identical, satisfying the requirement for perfect (provably secure) steganography. Any non-zero value implies a statistical deviation that could be exploited by an adversary.

    \item \textbf{Text Quality (PPL ↓)}: The fluency and naturalness of the generated text are assessed using perplexity (PPL). Perplexity measures how well the language model's probability distribution predicts the generated text. A lower PPL suggests that the stegotext is closer to what the model would naturally produce, indicating less degradation in text quality.

    \item \textbf{Computational Efficiency (Tok/Call ↑)}: To specifically measure the computational overhead introduced by our look-ahead mechanism, we define the metric Tokens per LLM Call (Tok/Call). This is the ratio of the total number of tokens in the final sequence to the total number of forward passes through the language model. A value of 1.0 corresponds to traditional single-pass generation methods, while a value less than 1.0 quantifies the additional computational cost of the recursive calls in our algorithm. We adopt this metric instead of wall-clock time to provide a more robust and hardware-independent measure of efficiency. Since factors such as the specific model, prompt, and GPU hardware can significantly affect execution time, and because all disambiguation methods share a common baseline cost (one LLM call plus ambiguity detection), Tok/Call directly isolates the core computational trade-off of our look-ahead strategy, which is directly proportional to its real-world runtime.
\end{itemize}


\begin{table*}[!t]
\centering
\caption{QUANTITATIVE COMPARISON OF \method{} WITH SYNCPOOL AND MWIS ON THE ENGLISH BENCHMARK (LLAMA3-8B/IMDB)}
\label{tab:english_results_merged}
\renewcommand{\arraystretch}{1.1} 

\subfloat[Results with iMEC encoder.\label{tab:eng_imec_sub}]{
    \footnotesize
    \setlength{\tabcolsep}{4pt} 
    \begin{tabular}{lccccc}
    \toprule
    \textbf{Method} & \textbf{top-$k$} & \textbf{BPT}~(↑) & \textbf{Tok/Call}~(↑) & \textbf{PPL}~(↓) & \textbf{KL}~(↓) \\
    \midrule
    Baseline & 16 & 1.51 & 1.00 & 7.13 & 0.00 \\
             & 32 & 1.86 & 1.00 & 9.18 & 0.00 \\
             & 128 & 2.33 & 1.00 & 12.90 & 0.00 \\
    \cmidrule{1-6}
    MWIS\cite{yan2023secure} & 16 & 1.59 & 1.00 & 8.80 & 1.80 \\
         & 32 & 1.75 & 1.00 & 12.93 & 2.70 \\
         & 128 & 2.24 & 1.00 & 27.44 & 5.91 \\
    \cmidrule{1-6}
    SyncPool\cite{qi2024provably} & 16 & 1.30 & 1.00 & 6.81 & 0.00 \\
             & 32 & 1.39 & 1.00 & 8.11 & 0.00 \\
             & 128 & 0.87 & 1.00 & 12.15 & 0.00 \\
    \cmidrule{1-6}
    \textbf{\method{}} & 16 & \textbf{1.49} & 0.63 & 6.89 & 0.00 \\
                       & 32 & \textbf{1.83} & 0.59 & 9.00 & 0.00 \\
                       & 128 & \textbf{2.32} & 0.42 & 12.68 & 0.00 \\
    \bottomrule
    \end{tabular}
}
\hfill 
\subfloat[Results with Discop encoder.\label{tab:eng_discop_sub}]{
    \footnotesize
    \setlength{\tabcolsep}{4pt} 
    \begin{tabular}{lccccc}
    \toprule
    \textbf{Method} & \textbf{top-$k$} & \textbf{BPT}~(↑) & \textbf{Tok/Call}~(↑) & \textbf{PPL}~(↓) & \textbf{KL}~(↓) \\
    \midrule
    Baseline & 16 & 1.72 & 1.00 & 6.84 & 0.00 \\
             & 32 & 2.14 & 1.00 & 8.64 & 0.00 \\
             & 128 & 2.82 & 1.00 & 12.12 & 0.00 \\
    \cmidrule{1-6}
    MWIS\cite{yan2023secure} & 16 & 1.75 & 1.00 & 8.77 & 1.79 \\
         & 32 & 2.25 & 1.00 & 12.35 & 2.75 \\
         & 128 & 2.92 & 1.00 & 28.94 & 6.25 \\
    \cmidrule{1-6}
    SyncPool\cite{qi2024provably} & 16 & 1.36 & 1.00 & 6.42 & 0.00 \\
             & 32 & 1.63 & 1.00 & 8.90 & 0.00 \\
             & 128 & 0.95 & 1.00 & 12.16 & 0.00 \\
    \cmidrule{1-6}
    \textbf{\method{}} & 16 & \textbf{1.67} & 0.63 & 6.95 & 0.00 \\
                       & 32 & \textbf{2.07} & 0.58 & 8.71 & 0.00 \\
                       & 128 & \textbf{2.66} & 0.42 & 12.82 & 0.00 \\
    \bottomrule
    \end{tabular}
}
\end{table*}

\begin{table*}[!t]
\centering
\caption{QUANTITATIVE COMPARISON OF \method{} WITH SYNCPOOL AND MWIS ON THE CHINESE BENCHMARK (QWEN2.5-7B/DOUBAN)}
\label{tab:chinese_results_merged}
\renewcommand{\arraystretch}{1.1} 

\subfloat[Results with iMEC encoder.\label{tab:chi_imec_sub}]{
    \footnotesize
    \setlength{\tabcolsep}{4pt} 
    \begin{tabular}{lccccc}
    \toprule
    \textbf{Method} & \textbf{top-$k$} & \textbf{BPT}~(↑) & \textbf{Tok/Call}~(↑) & \textbf{PPL}~(↓) & \textbf{KL}~(↓) \\
    \midrule
    Baseline & 16 & 1.56 & 1.00 & 4.39 & 0.00 \\
             & 32 & 1.74 & 1.00 & 5.44 & 0.00 \\
             & 128 & 2.09 & 1.00 & 7.85 & 0.00 \\
    \cmidrule{1-6}
    MWIS\cite{yan2023secure} & 16 & 1.19 & 1.00 & 4.02 & 2.21 \\
         & 32 & 1.47 & 1.00 & 4.90 & 2.75 \\
         & 128 & 1.75 & 1.00 & 7.00 & 2.97 \\
    \cmidrule{1-6}
    SyncPool\cite{qi2024provably} & 16 & 1.21 & 1.00 & 4.44 & 0.00 \\
             & 32 & 1.43 & 1.00 & 5.53 & 0.00 \\
             & 128 & 1.61 & 1.00 & 7.42 & 0.00 \\
    \cmidrule{1-6}
    \textbf{\method{}} & 16 & \textbf{1.53} & 0.75 & 4.47 & 0.00 \\
                       & 32 & \textbf{1.74} & 0.71 & 5.49 & 0.00 \\
                       & 128 & \textbf{2.07} & 0.66 & 7.72 & 0.00 \\
    \bottomrule
    \end{tabular}
}
\hfill 
\subfloat[Results with Discop encoder.\label{tab:chi_discop_sub}]{
    \footnotesize
    \setlength{\tabcolsep}{4pt} 
    \begin{tabular}{lccccc}
    \toprule
    \textbf{Method} & \textbf{top-$k$} & \textbf{BPT}~(↑) & \textbf{Tok/Call}~(↑) & \textbf{PPL}~(↓) & \textbf{KL}~(↓) \\
    \midrule
    Baseline & 16 & 1.66 & 1.00 & 4.31 & 0.00 \\
             & 32 & 1.97 & 1.00 & 5.40 & 0.00 \\
             & 128 & 2.53 & 1.00 & 7.67 & 0.00 \\
    \cmidrule{1-6}
    MWIS\cite{yan2023secure} & 16 & 1.30 & 1.00 & 3.92 & 2.25 \\
         & 32 & 1.61 & 1.00 & 5.02 & 2.57 \\
         & 128 & 2.17 & 1.00 & 7.02 & 3.03 \\
    \cmidrule{1-6}
    SyncPool\cite{qi2024provably} & 16 & 1.27 & 1.00 & 4.29 & 0.00 \\
             & 32 & 1.44 & 1.00 & 5.47 & 0.00 \\
             & 128 & 1.88 & 1.00 & 7.29 & 0.00 \\
    \cmidrule{1-6}
    \textbf{\method{}} & 16 & \textbf{1.57} & 0.75 & 4.37 & 0.00 \\
                       & 32 & \textbf{1.85} & 0.70 & 5.24 & 0.00 \\
                       & 128 & \textbf{2.38} & 0.65 & 7.72 & 0.00 \\
    \bottomrule
    \end{tabular}
}
\end{table*}

\begin{figure*}[!t]
\centering
\subfloat[Llama3 on IMDB (English)\label{fig:jsd_llama}]{
    \includegraphics[width=0.48\textwidth]{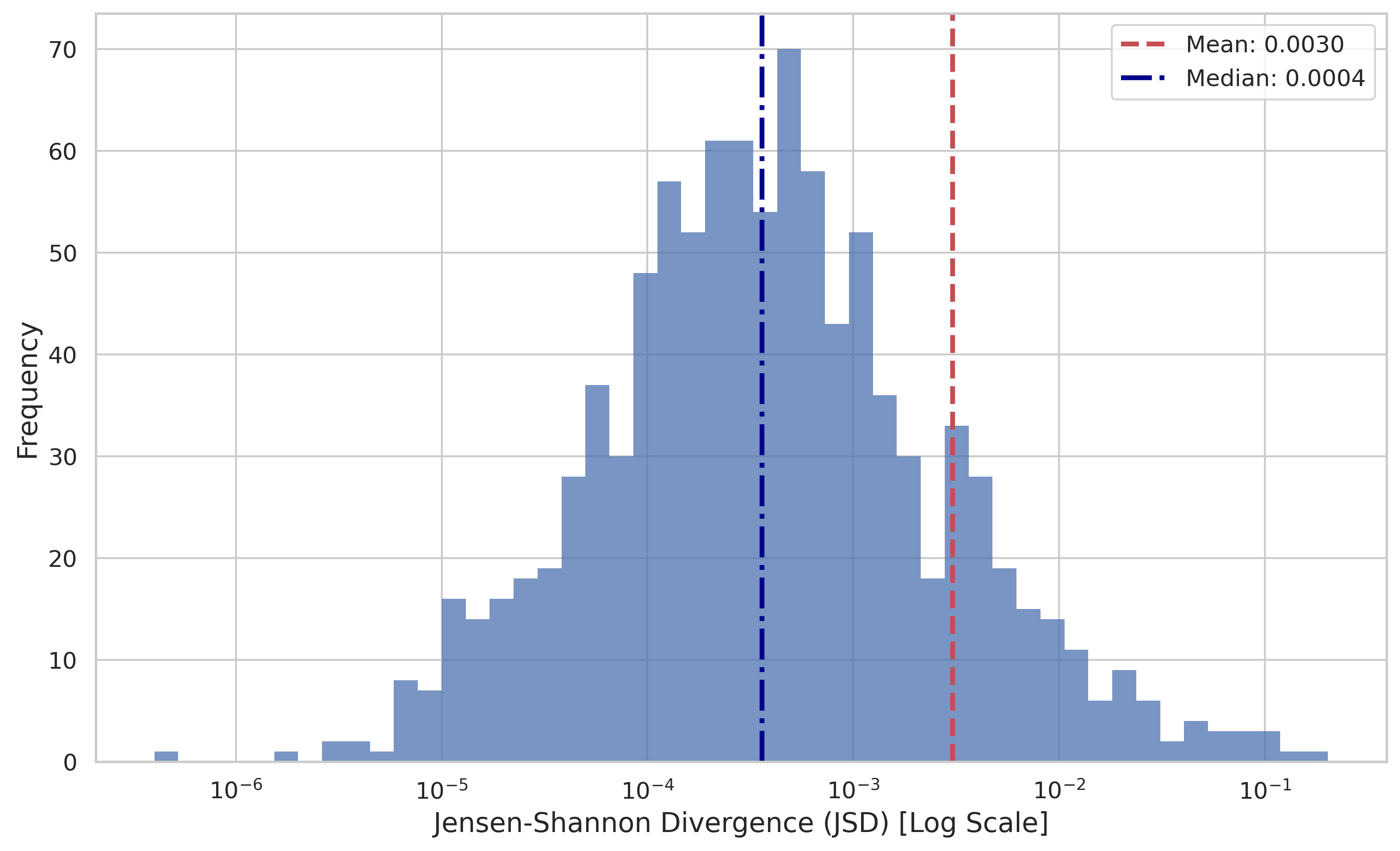} 
}
\hfill
\subfloat[Qwen on Douban (Chinese)\label{fig:jsd_qwen}]{
    \includegraphics[width=0.48\textwidth]{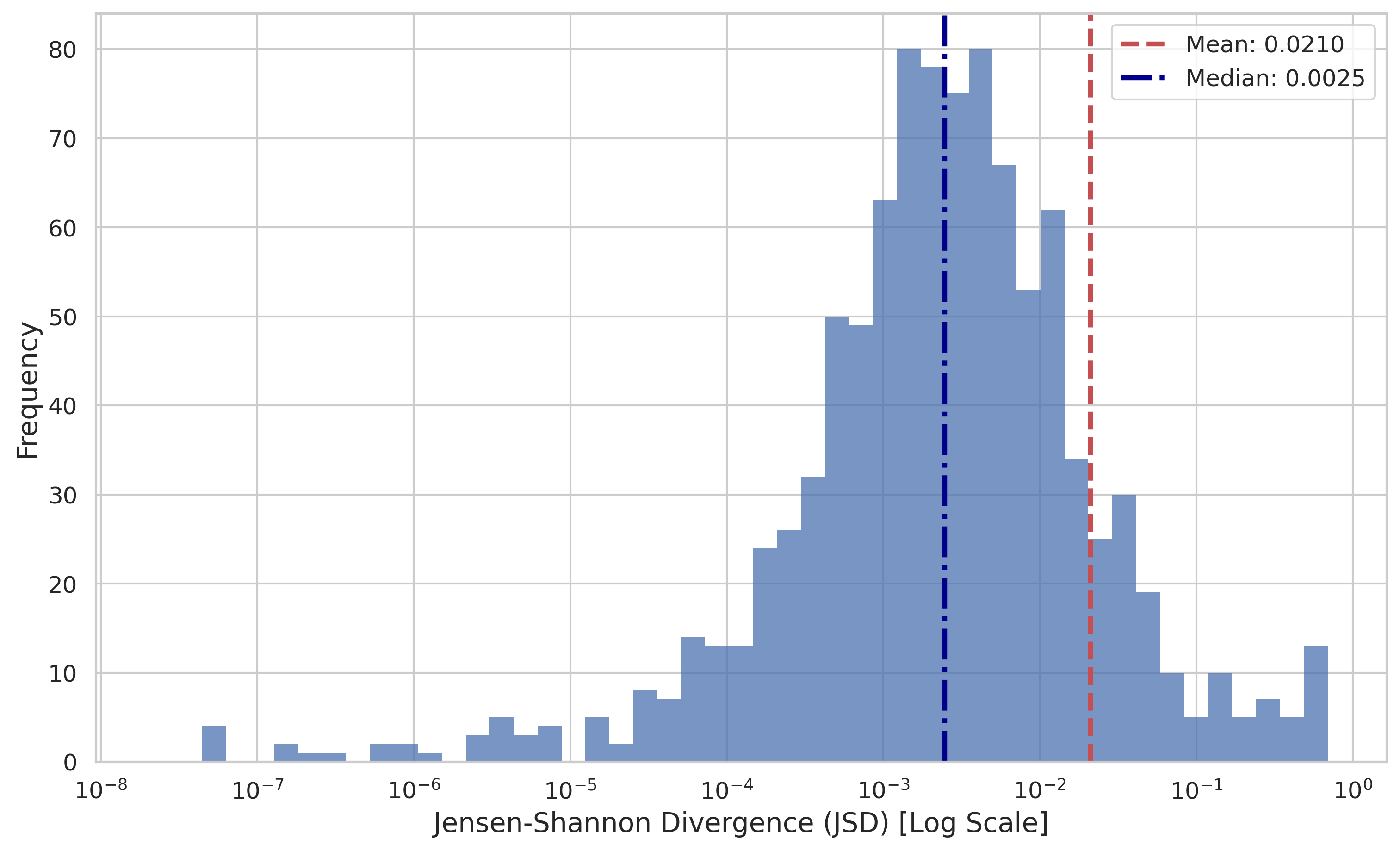}
}
\caption{Empirical distribution of Jensen-Shannon Divergence (JSD) scores for ambiguity scenarios on (a) Llama3 and (b) Qwen.}
\label{fig:jsd_distributions}
\end{figure*}

\subsection{Results and Analysis}
\label{subsec:results_analysis}

To empirically evaluate the performance of \method{}, we benchmark it against a carefully selected set of baselines across both English and Chinese tasks. The comprehensive results are presented in Tables~\ref{tab:english_results_merged} and~\ref{tab:chinese_results_merged}. In these tables, \textbf{Baseline} represents the direct application of the entropy encoder to the raw model output, serving as a theoretical capacity upper bound. The \textbf{MWIS} method denotes the state-of-the-art distribution-altering approach from Yan et al.~\cite{yan2023secure}, while \textbf{SyncPool} is the current state-of-the-art provably secure disambiguation algorithm from Qi et al.~\cite{qi2024provably}. Finally,\textbf{\method{}} refers to our proposed algorithm. This setup enables a direct and comprehensive comparison of capacity, security, and efficiency across the spectrum of existing approaches.

\subsubsection{Security}

A steganographic method's claim to be provably secure rests on its ability to maintain a probability distribution identical to the original cover source, which is quantified by a zero Kullback-Leibler (KL) divergence. Existing ambiguity resolution methods that rely on pruning candidates, such as \mbox{MWIS}, fundamentally disrupt this property. By altering the candidate pool, they inevitably introduce a non-zero KL divergence (up to 6.25 in our experiments), thereby forfeiting the guarantee of provable security. In contrast, as shown in Tables~\ref{tab:english_results_merged} and~\ref{tab:chinese_results_merged}, \method{} consistently maintains a KL divergence of 0.00. This confirms that its look-ahead mechanism is a distribution-preserving transformation, making it fully compatible with the strict requirements of a provably secure framework.

In terms of text quality, perplexity (PPL) measures the fluency of the generated stegotext. However, the ideal PPL is not necessarily the lowest possible value, but rather one that closely matches the PPL of the Baseline method. The Baseline reflects the statistical properties of text generated via standard random sampling from the model's true distribution. Any significant deviation from this PPL, whether higher or lower, suggests an unnatural generation process. The tables show that while \mbox{MWIS} can sometimes yield a lower PPL, it does so by altering the underlying probabilities. \method{}, on the other hand, produces PPL values that are consistently and remarkably close to the Baseline across all settings. This indicates that the text generated by \method{} is statistically indistinguishable in quality and naturalness from that of an unmodified language model, making it perceptually secure.

\subsubsection{Embedding Capacity}

A key contribution of this work is the high embedding capacity of \method{}, measured in bits Per Token (BPT). The method is designed to overcome the structural limitations of prior secure disambiguation techniques. As shown in Tables~\ref{tab:english_results_merged} and~\ref{tab:chinese_results_merged}, the BPT of \method{} scales effectively with the candidate pool size ($k$), closely following the trend of the theoretical maximum defined by the Baseline.

This scaling behavior differs from that of SyncPool. The capacity of SyncPool diminishes as $k$ increases, a consequence of its coarse-grained synchronization mechanism which discards the entire entropy of each selected ambiguity pool. A larger candidate pool size leads to more frequent and larger ambiguity pools, thus amplifying this entropy loss. For example, in the English/Discop experiment at $k=128$ (Table~\ref{tab:eng_discop_sub}), \method{} achieves a BPT of 2.66, which is a 180\% increase over SyncPool's 0.95 BPT.

\begin{figure}[!t] 
\centering 
\includegraphics[width=\columnwidth]{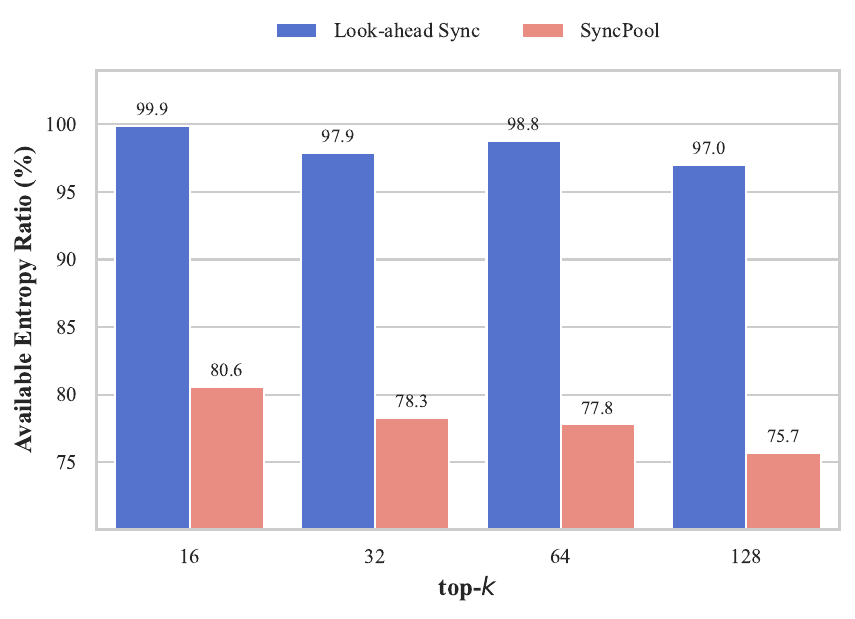} 
\caption{Comparison of the Available Entropy Ratio for different methods on the Qwen model as top-$k$ varies.} 
\label{fig:entropy_ratio} 
\end{figure}

The underlying mechanism for this performance is empirically illustrated in Figure~\ref{fig:entropy_ratio}, which plots the ratio of Shannon entropy available for encoding. The available entropy for \method{} remains above 97\% and is stable across different values of $k$. In contrast, the ratio for SyncPool drops below 76\% as ambiguity becomes more prevalent. This confirms that \method{}'s look-ahead mechanism successfully preserves and utilizes the entropy of non-terminal ambiguous paths, converting what would otherwise be lost information into usable embedding capacity.

The ability of \method{} to achieve a capacity close to the theoretical limit is further explained by the JSD distributions shown in Figure~\ref{fig:jsd_distributions}. The low Jensen-Shannon Divergence (JSD) scores (median < 0.003) between the future conditional distributions of ambiguous sequences indicate that the information loss from the synchronized sampling step is minimal in practice. This allows the computationally tractable, local resolution strategy to achieve a high capacity that approaches the theoretical capacity limit.

\subsubsection{Computational Efficiency}

The look-ahead mechanism that enables the high capacity of \method{} introduces additional computational overhead. This is quantified by the Tokens per LLM Call (Tok/Call) metric, which measures the average number of final tokens generated per forward pass of the language model. As shown in Tables~\ref{tab:english_results_merged} and~\ref{tab:chinese_results_merged}, the Tok/Call for \method{} is less than 1.0. This is a direct result of the additional LLM calls that are triggered when ambiguity is encountered in the candidate pool.

This computational overhead is a key characteristic of the method. In steganographic applications where maximizing embedding capacity is the primary objective, \method{} makes it possible to achieve substantial BPT gains, and this is associated with an increase in the number of LLM calls. The extent of this overhead is influenced by the sampling parameter $k$; larger values tend to increase the frequency of ambiguity, leading to a lower Tok/Call.

\section{Discussion}
\label{sec:discussion}

While the \method{} algorithm successfully enhances embedding capacity under a provably secure framework, two primary limitations warrant discussion. The first is the computational overhead introduced by the look-ahead mechanism, and the second is that the embedding capacity, while significantly improved, does not yet reach the theoretical upper bound established in our analysis.

The computational overhead stems from the additional language model forward passes required to execute the look-ahead strategy. This increase in computation is a direct trade-off for the algorithm's capacity gains, as each additional call serves to preserve and transform the Shannon entropy that is otherwise discarded by prior secure methods. For steganographic applications where achieving high embedding capacity is the principal objective, such as embedding substantial payloads within concise covertexts, this increased demand on computational resources may be an acceptable compromise.

The second limitation, the residual gap to the theoretical capacity limit, originates from the information loss inherent in the non-payload-bearing SyncSample step. A clear avenue for future research to mitigate this loss lies in the development of adaptive look-ahead strategies. A more sophisticated implementation could employ a heuristic, such as the entropy of the prefix set, $H(\mathcal{S}_{\text{prefix}})$, to dynamically guide its operations. For instance, a system could trigger a look-ahead call only when the potential capacity gain is significant, or even perform a more resource-intensive multi-path expansion for particularly high-entropy prefix sets. Such an approach could lead to a more optimal allocation of computational resources, further closing the gap to the theoretical capacity limit and enhancing the practical utility of the look-ahead paradigm.

\section{Conclusion}
\label{sec:conclusion}

A critical barrier to the practical use of provably secure linguistic steganography has been the substantial embedding capacity sacrificed by existing secure algorithms. In this paper, we identify the root of this limitation in the state-of-the-art method, SyncPool, whose coarse-grained synchronization mechanism systematically discards valuable Shannon entropy. To overcome this critical bottleneck, we introduce \method{}, a novel, recursive disambiguation algorithm. The core of our method is a look-ahead resolution strategy that performs minimal synchronized sampling only on truly indistinguishable token sequences.

By strategically preserving all other discernible candidate paths and reallocating their probability mass, \method{} maintains the rigorous zero-KL divergence guarantee of the synchronization paradigm while systematically eliminating the capacity constraints of prior work. Our approach ensures a uniquely determined decoding result for the receiver without forfeiting the informational content of distinguishable future paths.

Extensive experiments conducted on both English and Chinese language models validate our approach. The results demonstrate that \method{}'s embedding rate consistently approaches the theoretical capacity upper bound and substantially outperforms existing secure methods, with capacity gains exceeding 160\% in some settings. These gains are achieved without compromising the strict zero-KL security standard or the perceptual quality of the generated text. This work represents a significant step toward making high-capacity, provably secure linguistic steganography a practical and viable technology for covert communication.

\bibliographystyle{IEEEtran}
\bibliography{IEEEabrv,references}

%

\end{document}